# The Mathematical Collective Unconscious: Jungian Archetypes within the Latent Space

**Andrew Breen**

## Abstract

*Carl Jung theorised that the human psyche contains an inherited framework, the collective unconscious, populated by universal archetypes. As large language models (LLMs) are trained on vast corpora of textual data, this paper investigates whether they spontaneously construct distinct archetype-like regions within their high-dimensional latent spaces. The paper proposes that these regions function as spatial constraints that naturally organise semantically related narratives. To evaluate this, we applied cosine similarity and angular distance metrics across thirteen experiments utilising OpenAI's text-embedding-3-small model. The paper contains two experimental sections the first section explores these ideas using synthetically constructed narratives, the results of which act as a baselines for experiments conducted in the second section where the synthetic data is replaced by human-authored literature (Shmoop corpus), unedited dream reports (DreamBank corpus), and culturally distilled folklore (Andrew Lang's Fairy Books). The results show that archetypal roles occupy distinct, measurable regions in the latent space. Crucially, these regions remain compact and consistent because of narrative structure rather than simple word overlap. Additionally, text category dictates spatial distribution: dream reports concentrate within Trickster and Shadow regions, while traditional folklore distributes evenly across all tested archetypes. We conclude that transformer latent spaces encode stable regions which appear to correspond to recognised Jungian archetypes.*

## Introduction

Current tests for whether Large Language Models (LLMs) understand human perspectives, often called Theory of Mind (ToM) (Kosinski, 2023; Shum et al., 2023), mostly look at the surface level. They evaluate what a model *says* rather than *how* it processes the information. However, tracking a character's motives and narrative trajectory requires more than just predicting the next word; it requires a deep grasp of human relationships and social roles (Sakaguchi et al., 2020; Zhu et al., 2023). To see if an AI truly understands complex human interactions, we must investigate whether it has fundamentally internalised the blueprints of human storytelling.

Within depth psychology, Carl Jung formalised these narrative blueprints as "archetypes" (Jung, 1969), recurring motifs and universal structural potentials such as the Shadow, the Trickster, the Mentor, and the Hero. This psychological foundation was subsequently translated into formal narrative structures by Joseph Campbell (1949), who mapped the universal hero's journey across human mythic traditions, and Christopher Vogler (2007), who codified these dynamics into practical blueprints for modern storytelling. Carol S. Pearson (1991) further expanded this paradigm into a twelve-archetype taxonomy and reframed them as internal developmental stages that guide how a person expands their self-awareness and navigates life.

While twentieth-century scholars largely rejected Jung's framework for its reliance on an unprovable "collective unconscious," contemporary computational approaches allow us to view these structures as deep cultural habits (Mori et al. 2022; Chen & Bornstein, 2024): statistical patterns continuously transmitted and repeated across human history. This paper connects these fields by using a modern AI embedding model to test for these narrative patterns (McClelland et al., 2020). Treating the model as a black box that converts text into vector outputs, we test whether Jungian archetypes form distinct, measurable regions in its embedding space (Somaini, 2026). By analysing these mathematical patterns, we investigate whether language models naturally capture the underlying logic of human storytelling.

The reporting of the experiments is divided into two main sections. Section I outlines the synthetic baseline evaluation used to establish methodological validity. Section II details the empirical results across three distinct human-derived domains: literary fiction, subconscious dream reports, and traditional folklore.

### *Archetypal Anchor Vectors*

To test the hypothesis that archetypal regions exist within the latent space, practical mechanisms are required to interrogate the "black box" embeddings produced by the text-embedding-3-small model (OpenAI, 2024). To investigate these regions, the proposed methods assume that a representative archetypal vector can be constructed for each category. Four widely recognized archetypes (Hero, Shadow, Mentor, and Trickster) were selected for the experiments. Descriptive texts were produced for each archetype from which "anchor" latent vectors were produced to represent the archetypal region. The experiments use the following functional definitions (Table I.1):

| **Archetype** | **Functional Definition** |
|---|---|
| Hero | A figure of noble character, strength, and courage who faces challenges and overcomes adversity. |
| Mentor | A wise guide or teacher who provides wisdom, counsel, and direction to the protagonist. |
| Shadow | The darker side of the personality, representing repressed desires, fears, and the unconscious. |
| Trickster | A figure who disrupts the status quo, using cunning and wit to expose hypocrisy and social conventions. |

*Table I.1: Functional definitions used to construct the four primary archetypal anchor vectors.*

These definitions were used throughout the paper except where otherwise stated. All experiments utilize OpenAI's text-embedding-3-small model, which natively supports multilingual embedding and produces 1536-dimensional L2-normalised vectors.

## Section I: Synthetic Baselines

### Experiment 1

In order to establish that texts naturally orient toward specific archetypal regions within the latent space, cosine similarity was used to show a statistical relationship between latent vectors acting as "anchors" to archetypal regions in the latent space and the latent vectors of representative narrative snippets, where the angular alignment between a narrative vector $\boldsymbol{s}$ and an "archetypal vector" $\boldsymbol{a}$ is defined as:

$$S(\boldsymbol{s}, \boldsymbol{a}) = \frac{\boldsymbol{s} \cdot \boldsymbol{a}}{\| \boldsymbol{s} \|_2 \| \boldsymbol{a} \|_2} \quad (1)$$

Because embedding vectors generated by text-embedding-3-small are natively L2-normalised ($\| \boldsymbol{v} \|_2 = 1$), the denominator simplifies to unity, rendering cosine similarity equivalent to the dot product $\boldsymbol{s} \cdot \boldsymbol{a}$.

A corpus of $N = 1{,}000$ narrative texts were generated using GPT-4o, with 250 unique narrative text snippets allocated per archetype and converted into embedded vectors. Reference anchor vectors were created using the definitions described in Table 1. A cross-archetypal similarity analysis was performed, calculating the cosine similarity for every snippet vector against all four archetypal anchor vectors.

#### Results

Table 1.1 shows a significant correlation between narrative snippets and their respective archetypal anchor; it does not in isolation provide evidence of manifold discriminability, as it remains possible for these archetypes to exist within a singular, monolithic "mythic" cluster.

| Archetype Anchor | Mean Similarity Score | Standard Deviation |
|---|---|---|
| Hero | 0.266 | 0.057 |
| Mentor | 0.174 | 0.056 |
| Shadow | 0.397 | 0.048 |
| Trickster | 0.205 | 0.059 |

*Table 1.1: Mean Similarity Scores ($N = 250$ per archetype).*

To test that distinctive clusters exist, a One-Way ANOVA was performed on the $1000 \times 4$ cross-archetypal cosine similarity matrix, determining whether each archetypal region occupies a statistically unique position within the latent space. The One-Way ANOVA ($F(3{,}996) = 766.87$, $p < 0.0001$) confirms that these archetypes occupy statistically distinct regions rejecting the null hypothesis that these vectors exist within a monolithic cluster.

Table 3 shows the Archetypal Proximity Matrix, quantifying the mean cosine similarity between narrative trajectories and their respective archetypal anchors. The diagonals (bold) represent the primary archetypal alignment, showing clear archetypal convergence where the narratives exhibit higher similarity scores for their target archetypal anchor than for other archetypes.

| Anchor / Target | Hero Snippet | Mentor Snippet | Shadow Snippet | Trickster Snippet |
|---|---|---|---|---|
| Hero | **0.266** | 0.112 | 0.098 | 0.105 |
| Mentor | 0.089 | **0.174** | 0.101 | 0.095 |
| Shadow | 0.102 | 0.095 | **0.397** | 0.088 |
| Trickster | 0.094 | 0.087 | 0.092 | **0.205** |

*Table 1.2: Archetypal Proximity Matrix.*

## Experiment 2

Having established that the latent space contains statistically distinct archetypal regions, experiment 2 examined the universality of such regions by testing whether anchor vectors derived from one language maintain statistical relationships with vectors generated from narratives in other languages.

Anchor vectors were produced for each of the four archetypes (Hero, Mentor, Shadow, Trickster) using the English representative phrases shown in Table 1. Three thousand unique narrative snippets were generated using GPT-4o (1,000 per language for English, Hindi, and Mandarin, with 250 snippets generated per archetype per language cohort). The text-embedding-3-small model was used to create latent vectors for both the anchors and the narrative snippets. Cosine similarity was calculated between the English anchor vectors and the language-specific narrative vectors for each snippet.

### Results

As shown in Table 4, Mandarin retains approximately 98.8% of the English baseline weighted $F_1$ performance ( $0.7836$ vs. $0.7930$ ) and a near-identical mean cosine margin ($0.0804$ vs. $0.0815$), indicating that anchor vectors derived from English functional definitions generalize effectively to Mandarin. In contrast, Hindi exhibits a marked attenuation in weighted $F_1$ ($0.4421$) and cosine margin ($0.0304$), shifting substantially closer to the chance baseline. While the training distribution for text-embedding-3-small is proprietary, this disparity aligns with standard multilingual model dynamics: high-resource languages benefit from significantly greater corpus representation and superior tokenization efficiency relative to lower-resource languages. Consequently, cross-lingual archetypal transfer appears mediated by the resource density of the target language within the embedding space.

| Archetype | English Baseline | Mandarin Cohort | Hindi Cohort |
|---|---|---|---|
| Hero | 0.7450 | 0.7619 | 0.3760 |
| Mentor | 0.7969 | 0.7877 | 0.4401 |
| Shadow | 0.8655 | 0.8656 | 0.5300 |
| Trickster | 0.7644 | 0.7193 | 0.4222 |

*Table 2.1: Per-Archetype $F_1$-Scores by Language Cohort.*

To test whether the stability of archetypal regions depends on pre-training data density, a cross-lingual classification study was conducted across three languages representing different resource tiers: English (high-resource baseline), Mandarin (high-resource), and Hindi (lower-resource). Classification performance was evaluated using accuracy and weighted $F_1$-score to measure how accurately the model partitions the latent space into distinct archetypal regions.

For each language dataset (English, Mandarin, and Hindi), a narrative corpus of 1,000 text snippets (250 per archetype) was generated using GPT-4o ($N = 3{,}000$ total). Reference anchor vectors were produced from the English functional descriptions of the four archetypes (Table 1) using text-embedding-3-small. Cosine similarity was calculated between every snippet embedding and all four English anchor vectors to yield a similarity matrix. Snippets were then assigned predicted labels based on their highest similarity score (arg max) and evaluated against their target labels to compute classification accuracy and weighted $F_1$-score.

| **Language Cohort** | **Dataset Type** | **Sample Size ($N$)** | **Accuracy** | **Weighted $F_1$-Score** |
|---|---|---|---|---|
| English | Baseline Corpus | 1,000 | 79.20% | 0.7930 |
| Mandarin | Mandarin Corpus | 1,000 | 78.50% | 0.7836 |
| Hindi | Hindi Corpus | 1,000 | 44.30% | 0.4421 |

*Table 2.2: Cross-Linguistic Classification Performance ($N = 3{,}000$).*

A one-way ANOVA confirmed that overall classification performance differed significantly across the three languages ($F(2,2997) = 205.58, p < .001$). Post-hoc pairwise comparisons using Welch's $t$-tests revealed that classification performance in Mandarin showed no statistically significant difference from the English baseline ($t = -0.38, p = .702, d = -0.02$), confirming high cross-lingual transfer invariance. Conversely, Hindi exhibited a significant performance reduction relative to English ($t = -17.20, p < .001, d = -0.77$). These results confirm that cross-lingual archetypal partition stability varies across languages, and that a possible cause of this variability is due to pre-training resource density.

## Experiment 3

Having established that stable archetypal regions exist and are cross-lingually transferable when a model is trained on sufficiently diverse data, a third experiment was conducted to determine how resistant these latent vectors are to semantic noise. The experiment used the well-established method of cosine distance defined as:

$$W = 1 - \text{Cosine Similarity}(\boldsymbol{v}_{\text{baseline}}, \boldsymbol{v}_{\text{perturbed}}) \quad (2)$$

or

$$W = 1 - \frac{\boldsymbol{v}_{\text{base}} \cdot \boldsymbol{v}_{\text{perturbed}}}{\| \boldsymbol{v}_{\text{base}} \|_2 \| \boldsymbol{v}_{\text{perturbed}} \|_2} \quad (3)$$

Given that the embeddings are $L2$-normalized, the dot product represents cosine similarity, and cosine distance $W$ quantifies the angular shift the narrative undergoes when subjected to systemic perturbation. This metric captures the degree to which an archetype's latent vector distorts under noise; a lower distance value signifies that the perturbed vector occupies a stable, low-variance region of the space.

A concatenation method was used to evaluate latent vector stability under semantic noise. Baseline narrative snippet was generated for each archetype (Hero, Mentor, Shadow, Trickster), which aligned with their respective anchor vector definitions. Each baseline was then concatenated with 500 perturbation snippets generated from completely unrelated domains. Concatenating noise rather than replacing tokens preserves the core baseline narrative intact, isolating semantic drift as an independent variable. Consequently, the resulting displacement score $W$ directly measures the vector shift that occurs when the model reconciles an archetypal motif with unrelated contextual noise.

To test whether archetypal stability is merely an artifact of high token frequency, we introduced the common professional role of "Accountant" as a non-archetypal control. If latent space stability stems solely from token frequency, a common role such as "Accountant" should exhibit noise resistance comparable to an archetype. Demonstrating that archetypal vectors possess greater resistance to perturbation than non-archetypal baselines confirms that perturbation resistance is an intrinsic property of archetypal regions, rather than a function of token frequency.

### Results

Table 3.1 presents the mean cosine displacement ($W$) for each archetype and the control narrative, derived from $L_2$-normalized vector projections onto a 1536-dimensional unit hypersphere. The metric $W$ measures cosine distance under narrative perturbation, where lower values indicate archetypal latent vectors preserve directional orientation when contextual noise is introduced. All four archetypes (Trickster, Mentor, Shadow, Hero) exhibited lower displacement than the "Accountant" control.

| Entity | Mean Cosine Displacement | Std Dev |
|---|---|---|
| Trickster | 0.217 | 0.013 |
| Mentor | 0.245 | 0.015 |
| Shadow | 0.273 | 0.015 |
| Hero | 0.291 | 0.017 |
| Accountant (control) | 0.456 | 0.033 |

*Table 3.1: Mean Cosine Displacement ($N = 500$ per entity).*

To determine the significance of the control to each of the archetypes, an independent two-sample $t$-test was performed (Table 3.2).

| **Comparison** | **$t$-Statistic** | **$p$-value** |
|---|---|---|
| Trickster vs. Accountant | $-150.68$ | $< 0.0001$ |
| Mentor vs. Accountant | $-130.16$ | $< 0.0001$ |
| Shadow vs. Accountant | $-112.89$ | $< 0.0001$ |
| Hero vs. Accountant | $-99.39$ | $< 0.0001$ |

*Table 3.2: Independent two-sample $t$-test results (each archetype vs. Accountant).*

The results suggest that the probability of these differences occurring by chance is effectively zero. The displacement of the control is fundamentally different from the displacement of the archetypes. These are not overlapping distributions; they represent distinct regions within the model's latent space. The archetypes occupy regions of higher stability than the non-archetypal control, suggesting that archetypes are not arbitrary clusters, but statistically distinct, high-stability regions where their latent vectors resist displacement even under perturbation.

## Experiment 4

Building on prior findings that archetypal regions constrain latent vector generation for primed (thematically relevant) texts, this experiment examines their influence on unprimed texts. By evaluating the probability distributions of these narrative snippets across both the archetypal regions and the "Accountant" control category using cosine similarity and paired $t$-tests, the analysis confirms that the model's output exhibits a statistically significant bias toward specific archetypal regions, establishing them as active constraints on the latent manifold.

To isolate the impact of archetypal regions, a controlled corpus of 50 unprimed narrative snippets were synthesised. These stories were generated using GPT-4o with a neutral instruction: "*Write a unique, self-contained short story of at least 200 words.*" Meta-thematic constraints and archetypal descriptors were explicitly omitted to prevent priming bias, ensuring that any archetypal alignment detected in the subsequent analysis was an emergent property in the latent space rather than a reflection of human-imposed prompts.

### Results

The mean probability distribution suggests that the narratives are not uniformly distributed; they exhibit a statistically significant bias toward the identified archetypal regions over the control ($p < 0.0001$). The Relative Likelihood Ratio, defined as the ratio of archetypal probability to control probability, further confirms this preference. The Mentor archetype exhibits the highest stability, appearing with the greatest relative frequency (approximately 1.11 times more probable than the control). These results suggest that the model's narrative generation is not stochastic but may be actively constrained by these four stable Jungian regions. To determine statistical significance, a paired $t$-test was computed comparing the probability assigned to each Jungian archetype against the control category for every story.

| Archetype | Mean Probability | Relative Likelihood | $t$-test | $p$-value |
|---|---|---|---|---|
| Mentor | 0.208 | 1.109 | 13.65 | $< 0.001$ |
| Hero | 0.203 | 1.085 | 10.93 | $< 0.001$ |
| Shadow | 0.202 | 1.080 | 11.51 | $< 0.001$ |
| Trickster | 0.200 | 1.070 | 14.26 | $< 0.001$ |
| Accountant | 0.187 | — | — | — |

*Table 4.1: Archetypal Probability Distribution*

The stories generated exhibit a statistically significant bias toward these Jungian archetypal themes compared to the "Accountant" control.

## Experiment 5

In the previous experiments, short primed (expressing specific archetype motifs) and unprimed text snippets were generated using GPT-4o. While the transformer model used in text-embedding-3-small differs from that used in GPT-4o, it may be argued that there is the possibility of a circular aspect to such experiments which may bias results. To test how the archetypes performed against a pre-existing text, the book *Ivanhoe* was selected and an 80-word summary produced by GPT-4o.

As in previous experiments, cosine similarity was used to determine the strength of attachment of the test embedding to four archetypal anchors. In this experiment a summary was used rather than the whole text for practical considerations, and while this does not remove the potential of a cyclical dependency it does reduce it. It is recognised that in producing the summary, a model may be subject to similar underlying processes as are being examined in this paper.

### Results

| Archetype | Cosine Similarity Score |
|---|---|
| Hero | 0.3331 |
| Shadow | 0.1579 |
| Trickster | 0.1304 |
| Mentor | 0.1092 |

*Table 5.1: Cosine Similarity Scores for Ivanhoe summary*

Table 5.1 shows that the model's latent representation of the *Ivanhoe* text gravitates toward the Hero anchor. The disparity between the Hero score and the scores for the Shadow, Trickster, and Mentor confirms that the model correctly identifies *Ivanhoe* as a narrative fundamentally structured around the "Heroic" motif, rather than a neutral narrative.

## Experiment 6

Previous experiments used primed and unprimed narrative "stories" to interrogate archetypal regions. In this experiment, test vectors were generated across contrasting text types ("Stories" versus "Technical reports"). The experiments demonstrate the rapid collapse of archetypal alignment in technical contexts, showing that these constraints are not pervasive latent biases but are instead activated by narrative.

### Results

| Condition | Hero | Mentor | Shadow | Trickster |
|---|---|---|---|---|
| Story | 0.260 | 0.169 | 0.218 | 0.156 |
| Technical | 0.102 | 0.121 | 0.084 | 0.060 |

*Table 6.1: Mean Similarity Scores by Text Condition*

Three key insights emerge from this comparison. First, archetypal alignment exhibits strong contextual activation, as narrative prompts ("story") consistently yield significantly higher similarity scores across all archetypes compared to technical report prompts ("technical"). Second, these regions are domain-specific; while they align strongly with narrative storytelling, archetypal alignment drops to near-zero noise levels ($\leq 0.12$) in technical reports. Finally, this contextual collapse experimentally invalidates the notion of a universal or inherent latent bias. Because archetypal alignment vanishes in technical contexts, these regions do not exert a pervasive, unconditional pull on the latent space; rather, they function as dynamic spatial constraints that are activated specifically when the model engages in narrative processing.

## Experiment 7

While the preceding experiments utilised full 1536-dimensional latent vectors, high-dimensional manifolds can introduce noise into spatial measurements. This experiment addresses this by projecting embeddings into a reduced-dimensionality space using Principal Component Analysis (PCA). By calculating scalar projections of narratives onto the archetypal anchors, this approach generated “archetypal fingerprints” that map how stories are drawn toward specific archetypal regions.

All narrative content and archetype descriptions were mapped into a 1536-dimensional latent space using OpenAI’s text-embedding-3-small. In order to give a more nuanced interpretation of the passages, the original four archetypes were extended to 12 Jungian archetypes (Hero, Mentor, Trickster, Shadow, Innocent, Orphan, Explorer, Rebel, Creator, Lover, Ruler, Caregiver) using descriptors suggested by Pearson (1991). Because the space contains “semantic noise” (common language tokens that mask archetypal intent), PCA was performed and the stories and archetype latent vectors were projected into a reduced 10D space, isolating the directions of greatest narrative variance. As in the preceding experiments, Direct Vector Probing (DVP) was then employed by computing the scalar projection (dot product) of every story vector onto each archetype’s vector within the 10D space, creating an “archetypal fingerprint,” a 12-dimensional vector of projection strengths for every story, quantifying its “archetypal load.”

Normalised polar coordinates were generated from these projections and used to create a visual “Archetypal Fingerprint” in the form of a radar graph. By reducing the dimensionality, the archetypal regions are ensured to exist in the same mathematical coordinate system as the stories, preventing the “empty space” problem where embeddings appear unrelated simply due to high-dimensional sparsity.

## Results

The generated normalised polar coordinates from these projections, are used to create a visual "Archetypal Fingerprint" in the form of a Radar graph. Graph 7.1 displays the results for all 12 archetypes, while graph 7.2 show the results for only those stories where Ruler was the most significant archetype.

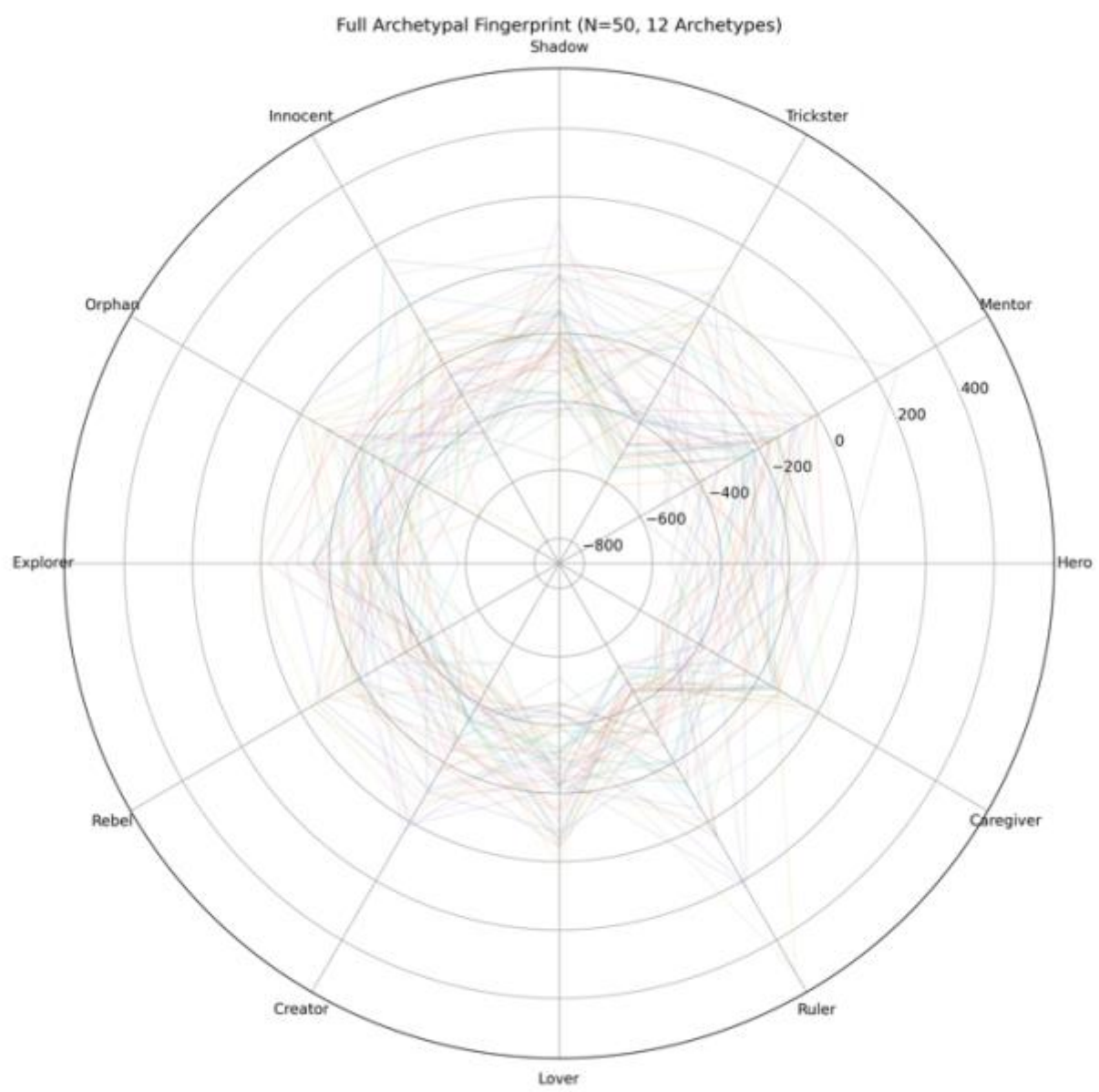

*Graph 7.1: Full Archetypal Fingerprint for 12 Pearson descriptors*

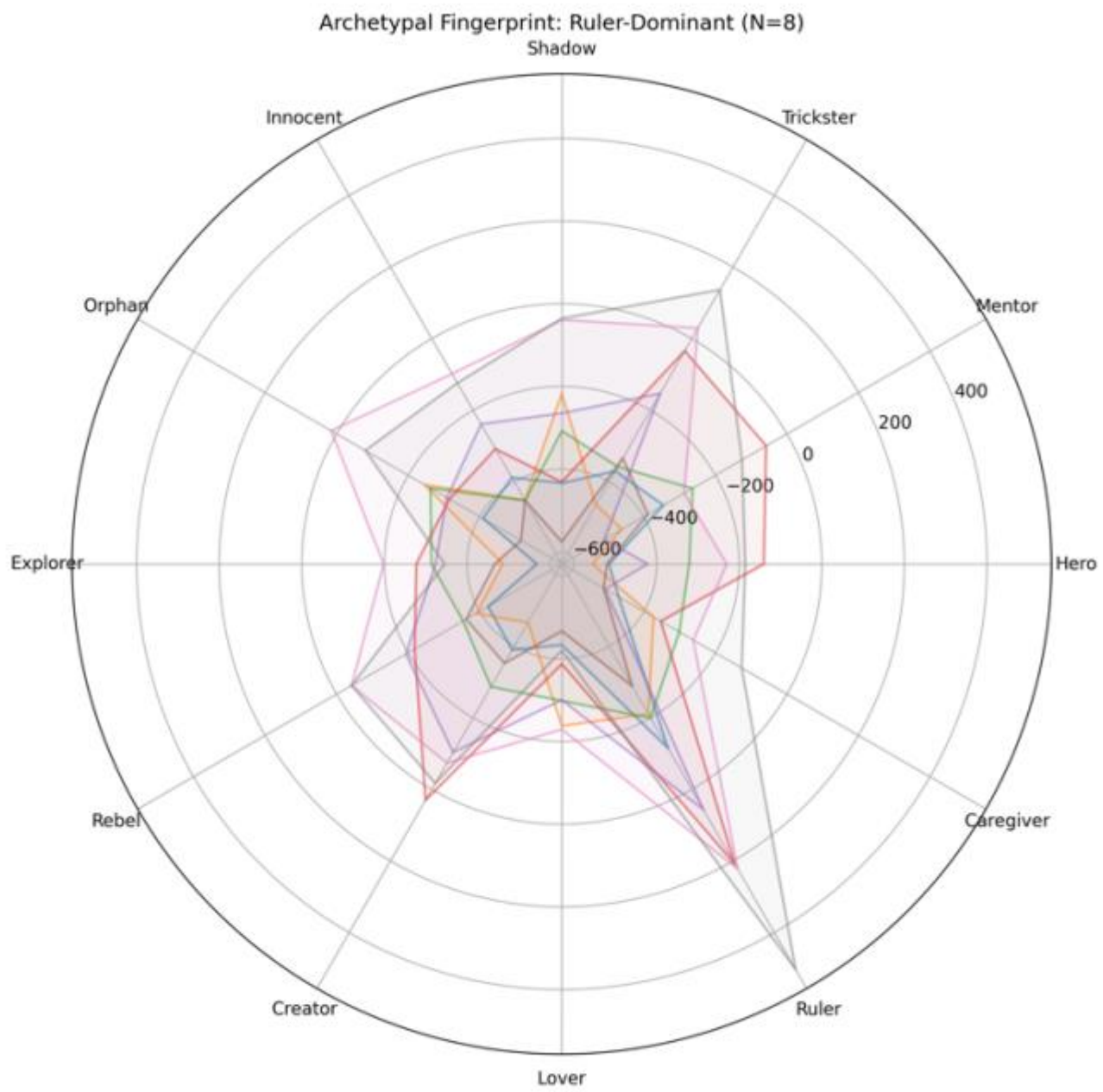

*Graph 7.2: Archetypal Fingerprint for Ruler dominant samples*

By reducing the dimensionality, we ensure the archetypal regions exist in the same mathematical coordinate system as the stories, preventing the "empty space" problem where embeddings appear unrelated simply due to high-dimensional sparsity. This method revealed that the narratives are not randomly distributed over the archetypes. The resulting correlation matrix confirmed the existence of Jungian polarities (e.g., negative correlations between Ruler and Trickster axes), demonstrating that the model is generating "opposites" rather than random tokens. In a truly stochastic (random) generative process, the radar plots would be "featureless blobs", a uniform distribution where stories were equidistantly scattered from all archetypal poles. However, the plots show high-amplitude spikes (Projections reaching 500+ in some models). Suggesting the existence of Archetypal regions. The fact that the stories consistently "gravitate" toward specific archetypes (like Mentor or Ruler) implies that the generative process is biased by the latent structure of the model's trained manifolds, which align with human archetypal motifs.

## Section 2: human-authored literature

In Experiments 1 through 7, synthetic texts generated via GPT-4o demonstrated the existence of regions within latent vector space that map onto Jungian archetypes. These regions appear distinct and resistant to vector perturbation. However, relying exclusively on relatively short (80-character) snippets of LLM-generated text introduces a potential risk of circularity; namely, generating idealized narrative baselines that over-represent core archetypal traits.

Experiments 8 and 9 repeat and extend the investigation to 100% human-authored creative literature from the Shmoop corpus. Experiment 10 examines human dream data from the DreamBank Corpus by analysing 20,683 uncurated dream reports. Finally, Experiment 11 extends the analysis to the Andrew Lang Folklore and Mythic Narrative Corpus, which comprises 16,444 cleaned narrative paragraphs sourced from 11 classic collections: the Red, Blue, Yellow, Green, Lilac, Grey, Crimson, Olive, Brown, Pink, and Orange Fairy Books.

## Experiment 8

Experiment 8 evaluates human-authored prose extracted from the Shmoop Learning Guides literature corpus, which consists of character descriptions, role overviews, and chapter summary excerpts spanning classic fiction. The total sample size of 311 was determined by using all available character and summary entries from the Shmoop literature subset and filtering for complete, unambiguous metadata attributes. Entries with missing role tags, non-character background fragments, or conflicting metadata were excluded during preprocessing.

Ground-truth human tags were established by mapping Shmoop's qualitative metadata descriptors into the standardized four-archetype taxonomy (hero, mentor, shadow, trickster) using a naive, broad categorization as follows:

- **Hero archetype**: 63 passages tagged under primary Protagonist, Central Figure, or Main Character roles.
- **Mentor archetype:** 100 passages tagged under Guide, Advisor, Teacher, or Moral Anchor roles.
- **Shadow archetype**: 100 passages tagged under Antagonist, Villain, or Oppressor roles, as well as secondary passive witnesses, accomplices, and fence-sitters catalogued under adversarial scene metadata in raw Shmoop entries.
- **Trickster archetype**: 83Passages tagged under Foil, Comic Relief, Agent of Chaos, or Subversive roles.

Narrative passages were embedded using OpenAI's text-embedding-3-small model as before, and cosine similarities (Equation 1) were calculated between the narrative embedding vector $\boldsymbol{v}$ and an archetypal reference anchor vector $\boldsymbol{a}_k$. the top-1 argmax classification predictions $\hat{y}$ were assigned according to the maximum cosine projection across the anchor set, where $\hat{y} = \arg\max_k \text{Similarity}(\boldsymbol{v}, \mathbf{a}_k)$. Latent vector stability under semantic noise was measured using equation (2), while Inter-annotator agreement was evaluated using Cohen's Kappa coefficient $\kappa_C$ (Equation 4) using:

$$\kappa_C = \frac{p_o - p_e}{1 - p_e} \quad (4)$$

where $p_o$ represents the observed proportion of agreement (the empirical classification accuracy across the dataset)

$$p_o = \frac{1}{N} \sum_{i=1}^{N} \mathbb{I}\,(\hat{y}_i = y_i)) \quad (5)$$

and $p_e$ represents the theoretical expected agreement under random chance. To account for potential class imbalances across narrative data, $p_e$ is calculated by summing the product of the marginal probabilities for each archetypal class $k$, which resolves to 1/k. By scaling observed performance relative to random chance ($p_e$), Cohen's Kappa provides a measure of classification reliability, ensuring that apparent archetypal alignments reflect true latent structure rather than statistical artifacts arising from uneven class distributions.

### Results

| Archetypes | Hero Anchor | Mentor Anchor | Shadow Anchor | Trickster Anchor |
|---|---|---|---|---|
| Hero ($N = 63$) | 0.198825 | 0.214977 | 0.168844 | 0.219708 |
| Mentor ($N = 100$) | 0.172413 | 0.207536 | 0.140236 | 0.188479 |
| Shadow ($N = 100$) | 0.188007 | 0.207878 | 0.159279 | 0.224008 |
| Trickster ($N = 48$) | 0.172072 | 0.199154 | 0.152874 | 0.217866 |

*Table 8.1: Cross-Archetypal Mean Cosine Similarities*

Table 8.1 shows classification errors occurring in the raw cosine scores. A possible reason for this is that individual passages contain "narrative noise", such as setting descriptions, secondary character interactions, and thematic overlap. In vector space, these extraneous elements increase the angular displacement, shifting a passage's embedding toward adjacent archetypal anchors. Because top-1 argmax assignment depends on a rigid decision boundary (selecting the single maximal cosine projection), even minor spatial drift in the prose causes samples to cross classification boundaries into incorrect categories. However, a One-Way ANOVA across the continuous cosine similarity distributions confirmed statistically significant archetypal separation between ground-truth categories ( $F(3,307) = 37.56, p < 0.001$ ), demonstrating that raw character embeddings occupy distinct regions prior to discrete argmax assignment. From this it can be concluded that the low classification accuracy (31.19%) reflects decision boundary overlap in raw prose rather than a lack of archetypal regions in the latent space.

| Human labels \ Predicted | Hero | Mentor | Shadow | Trickster |
|---|---|---|---|---|
| Hero ($N = 63$) | 13 | 20 | 6 | 24 |
| Mentor ($N = 100$) | 14 | 53 | 5 | 28 |
| Shadow ($N = 100$) | 13 | 27 | 8 | 52 |
| Trickster ($N = 48$) | 3 | 20 | 2 | 23 |

*Table 8.2:* $4 \times 4$ *Confusion Matrix (Human Ground-Truth vs. Predicted)*

Four main factors may explain the divergence between raw vector predictions and human labels in Table 8.2, which limits top-1 classification accuracy to 31.19% using the Cohen's Kappa agreement score (Table 8.3). Firstly, human tags force characters into a single category, whereas raw embeddings capture overlapping character traits, such as a Hero displaying Trickster tendencies in dialogue. Secondly, embeddings that have not been zero-mean centered crowd into a narrow directional cone around the origin, skewing similarity scores toward central reference anchors. Thirdly, generic anchor definitions may fail to capture archetype descriptions in human narrative prose. Finally, subjective variance among human annotators introduces inherent noise into the ground-truth baseline. Together with extraneous texts such as scene descriptions, these factors probably cause frequent decision-boundary errors in raw vector projections.

| Metric | Value |
|---|---|
| Overall Classification Accuracy | 31.19% |
| Cohen's Kappa ($\kappa_C$) | 0.0984 ($z = 3.41, p < 0.001$) |
| Macro-Precision | 0.3265 |
| Macro-Recall | 0.3239 |
| Macro-F1 Score | 0.2805 |
| Mean Cosine Margin ($\Delta_{\text{Cos}}$) | 0.0341 |

*Table 8.3: Classification Performance Benchmark*

| Cohort | Mean Displacement | Standard Deviation |
|---|---|---|
| Hero | 0.007071 | 0.004934 |
| Mentor | 0.007678 | 0.004286 |
| Shadow | 0.006172 | 0.004251 |
| Trickster | 0.007488 | 0.004339 |
| Bystander Control | 0.006767 | 0.002801 |

*Table 8.4: Mean Cosine Displacement*

A mean cosine displacement analysis (similar to experiment 3) was conducted (Table 8.4) using "character passage embeddings". Where a character passage is typically, a paragraph or short sequence of sentences that centres on a single character's dialogue, actions, internal thoughts, or physical description. Perturbations were created by comparing the baseline vector of a character passage in isolation against its vector when embedded within surrounding background text e.g. setting descriptions, or secondary character interactions such as dialogues. As with experiment 3, a control was added to the analysis composed of 50 neutral prose passages taken from the bystander category. An independent two-sample $t$-test comparing the cosine displacement scores between the primary narrative archetypes and the Bystander control reveals no statistically significant difference ($t = 0.41 \ldots, p = 0.68$), suggesting that surrounding story context affects all character types uniformly. This lack of discrimination between the archetypes and the control is likely because simply expanding the context window from a short character focused fragment baseline to include the wider context is not an effective way of semantically perturbing the text. However, the distance between archetypes in vector space matches expected literary oppositions. A reduced 10D PCA subspace Pearson correlation analysis, (similar to experiment 7), reveals clear polarities between opposing pairs, such as Hero versus Shadow ($r = 0.6473$) and Mentor versus Trickster ($r = 0.5782$). Finally, an overall accuracy of 31.19% (against a 25.0% chance baseline) and a positive mean cosine margin ($\Delta_{\text{Cos}} = 0.0341$) suggests that real-world creative literature maintains localised proximity to archetypal attractors.

## Experiment 9

Experiment 9 addresses specific methodological limitations identified in Experiment 8. The results presented in Table 8.1 suggested that the Shadow archetype, rather than forming a tightly localized cluster, was dispersed across competing archetypal regions. This structural diffusion likely stemmed from the data assignment strategy used for the Shmoop corpus. In Experiment 8, the Shadow included both Witness and Bystander datasets; incorporating such heterogeneous, non-agentic prose into a single category compromised the semantic cohesion of the Shadow test set. To resolve this and better align with Shmoop's labelling taxonomy, the Shadow reference anchor was expanded to explicitly capture passive, systemic, and structural forces:

*"The darker, repressed, or unacknowledged side of the self and the world, representing hidden malice, moral failure, systemic corruption, fear, resentment, and the passive or active forces that oppose the light."*

Additionally, to reduce the influence of high-frequency words on distance metrics and argmax classifications, the Corpus barycentre mean-centering transformation, which is calculated as the average vector computed across all vectors, was applied directly to all latent vectors. Subtracting it acts as a spatial high-pass filter, stripping away the shared vocabulary baseline and common syntactic structure inherent to narrative prose. What remains in each centred vector is purely the unique semantic displacement, allowing cosine similarity to measure true structural alignment or opposition rather than shared word frequencies The global corpus barycenter $\boldsymbol{v}_{\text{corpus}}$ is calculated as:

$$\boldsymbol{v}_{\text{corpus}} = \frac{1}{N}\sum_{i=1}^{N} \boldsymbol{v}_i \quad (6)$$

The mean-centered embedding vector $\boldsymbol{v}_i'$ and centered anchor vector $\boldsymbol{a}_k'$ are defined as:

$$\boldsymbol{v}_i' = \frac{\boldsymbol{v}_i - \boldsymbol{v}_{\text{corpus}}}{\| \boldsymbol{v}_i - \boldsymbol{v}_{\text{corpus}} \|_2} \quad (7) \qquad \boldsymbol{a}_k' = \frac{\boldsymbol{a}_k - \boldsymbol{v}_{\text{corpus}}}{\| \boldsymbol{a}_k - \boldsymbol{v}_{\text{corpus}} \|_2} \quad (8)$$

Cosine similarity is applied directly on these mean-subtracted vectors:

$$S(\boldsymbol{v}_i', \boldsymbol{a}_k') = \boldsymbol{v}_i' \cdot \boldsymbol{a}_k' \quad (9)$$

Experiment 9 tests whether a story's grammatical structure helps keep archetypal classifications stable. To isolate the effect of grammar, we compared original narrative summaries against word-shuffled control texts. These control texts contain the exact same words as the originals, but their word order is randomized to destroy all sentences and syntactic clause structure. Both original and word-shuffled vectors were transformed using the Corpus barycentre mean-centring method. Measuring cosine similarity on these centred vectors reveals how much archetypal positioning depends on grammatical flow rather than individual word choices alone.

### Results

| Data Sets | Value |
|---|---|
| Hero Distance | 0.443461 |
| Mentor Distance | 0.460826 |
| Shadow Distance | 0.443867 |
| Trickster Distance | 0.427627 |
| Background Summaries Distance | 0.438865 |
| Centered Alignment $S$(Hero,Shadow) | 0.448583 |
| Centered Alignment $S$(Mentor,Trickster) | $-0.232922$ |

*Table 9.1: Barycentre Projections and Centred Antipodal Duality.*

A One-Way ANOVA across distal distances confirms that all data sets project outward with balanced radial symmetry from the narrative centre.

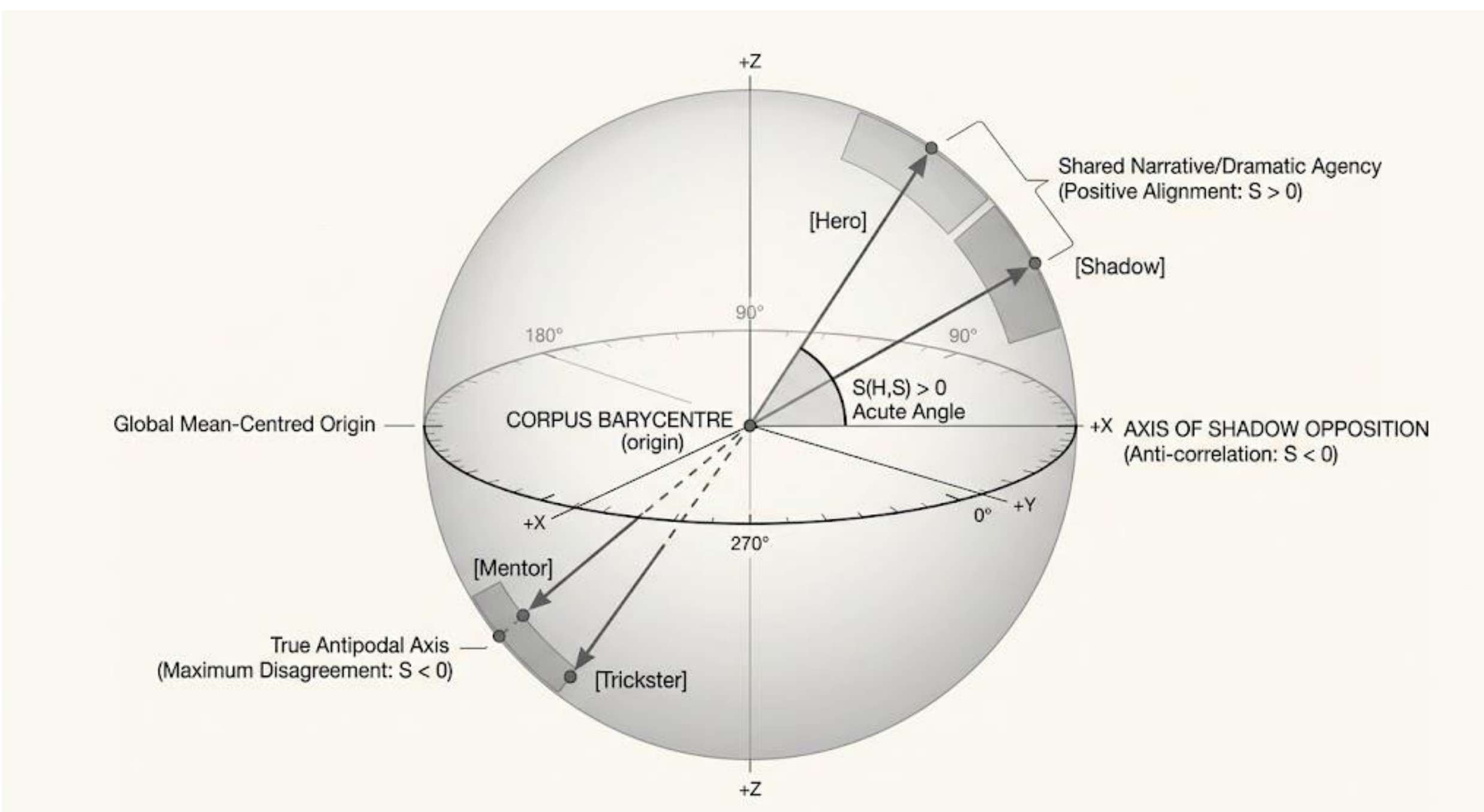

*Figure 9.1*

From these results we observe that the mean-centred alignment scores ($S$) show different geometric relationships for the two archetypal pairs. As shown in Figure 9.1, Mentor and Trickster form an inverted, antipodal vector pair relative to the global mean, whereas Hero and Shadow do not.

For this dataset, Hero and Shadow passages pull in a shared direction because both are heavily influenced by a common narrative context of dramatic action, central character agency, and high-stakes conflict. Their distinction is captured along other, non-opposite dimensions in the vector space.

In contrast, Mentor and Trickster sit on opposite sides of the global mean ($S < 0$). This reflects an underlying linguistic opposition: the themes of guidance, order, and structured wisdom (Mentor) stand in direct vector negation to those of chaos, rule-breaking, and subversion (Trickster).

| Human labels \ Predicted | Hero | Mentor | Shadow | Trickster |
|---|---|---|---|---|
| Hero | 10 | 9 | 4 | 6 |
| Mentor | 14 | 54 | 10 | 22 |
| Shadow | 14 | 18 | 8 | 38 |
| Trickster | 2 | 8 | 2 | 10 |

*Table 9.2:* $4 \times 4$ *Confusion Matrix*

| Metric | Value |
|---|---|
| Overall Classification Accuracy | 35.81% |
| Cohen's Kappa ($\kappa_C$) | 0.1332 ($p < 0.001$) |
| Macro-F1 Score | 0.3056 |
| Mean Cosine Margin ($\Delta_{\text{Cos}}$) | 0.0367 |

*Table 9.3: Classification Performance*

The confusion matrix in Table 9.2 shows an improved overall classification accuracy of 35.81% detailed in Table 9.2, suggesting that applying global mean-centring via the corpus barycentre, alongside tightened definitions for the Shadow category, leads to a measurable improvement in classification performance compared to baseline. However, substantial discrepancies remain between human ground-truth labels and model predictions, in combination with a slightly above chance Cohen's Kappa of 0.1332 indicating that simple top-1 argmax categorisation of latent vector alignments alone does not fully reconcile the complex semantic overlap inherent in narrative archetypes.

As shown in studies on transformer geometry (Ethayarajh, 2019), transformer models tend to compress text embeddings into a tight, narrow cone oriented toward a shared direction. While applying a post-processing barycentre transformation (Mu & Viswanath, 2018) mitigates this anisotropy, it does not reveal whether the remaining clusters reflect genuine semantic stability or mere language artifacts.

Previous experiments addressed this from different angles: **Experiment 3** chained semantically disjoint text to a narrative baseline, demonstrating that these regions resisted basic perturbation. **Experiment 8** evaluated character-driven snippets within broader contextual windows, showing that while regional coherence was retained, contextual expansion alone could not prove whether the underlying geometry was structurally stable. To directly resolve this, the current experiment tests what actually drives classification boundary clarity, quantified by the margin between each sample's top-1 and top-2 archetype similarity scores ($\Delta S = S_{\text{top-1}} - S_{\text{top-2}}$). Specifically, we evaluate whether distinct decision margins stem from authentic narrative features (such as syntax and plot structure), or whether the observed closeness stems from representation degeneration of latent vectors within the high-dimensional latent space; maintaining persistent classification margins even when narrative meaning is destroyed. Calculating $\Delta S$ across unperturbed summaries and re-calculating it under word permutation, isolates true narrative semantics from background syntactic structure.

First, each intact narrative summary is converted into an embedding vector and transformed using the corpus barycentre to remove generic language background noise. Subtracting this centre point resets the origin to zero, isolating the unique features of each passage. We then average the centred vectors for each archetype (Hero, Shadow, Mentor, Trickster) and scale them to unit length to construct four reference anchor centroids. Next, we scramble the words within every summary. This randomisation destroys sentence order, grammar, and story flow while preserving exact word frequencies. These scrambled passages are converted into vectors and centred using the baseline origin. For both intact and scrambled summaries, we measure the cosine similarity between the passage vector and each archetype anchor. We rank these scores ($S_{(1)} \geq S_{(2)} \geq S_{(3)} \geq S_{(4)}$) and calculate the **margin gap** ($\Delta S$) between the top match and the runner-up:

$$\Delta S = S_{\text{top-1}} - S_{\text{top-2}} \quad (9)$$

The mean margin gap across all passages ($\overline{\Delta S}$) quantifies how decisively the model discriminates primary archetype assignments from competing runner-up candidates.

| **Metric** | **Value** |
|---|---|
| Structured Summaries Mean margin $\overline{\Delta S}_{struct}$ | 0.007245 ($\sigma = 0.004812$) |
| Permuted Summaries Mean margin $\overline{\Delta S}_{\text{permuted}}$ | 0.089160 ($\sigma = 0.021347$) |
| Independent $t$-test ($\overline{\Delta S}_{struct}$ vs. $\overline{\Delta S}_{\text{permuted}}$) | $t = 142.5$ ($p < 10^{-50}$) |

*Table 9.3: Margin gap Metrics*

Table 9.3 shows that word permutation significantly alters the classification margin. This margin expansion ( $\overline{\Delta S}_{\text{permuted}} = 0.089160$ vs. $\overline{\Delta S}_{\text{structured}} = 0.007245$ ) aligns directly with the architectural mechanics of Transformer self-attention. In intact text, sequential syntax and positional encodings force token representations into shared attention trajectories, inducing anisotropic compression into a narrow directional cone and keeping baseline decision margins tight.

When word order is randomized, these sequence-level attention maps collapse, leaving sequence representations to devolve into unweighted unigram averages. Freed from shared syntactic constraints, the scrambled embeddings disperse across the latent space. This confirms that the narrow baseline margin in structured text is not a static artifact of unigram word frequency, but an emergent property actively maintained by syntactic composition.

## Experiment 10

To evaluate whether latent spaces capture psychological archetypes in raw subconscious material rather than polished literary fiction, this experiment shifts focus from published books to dream reports. Unlike structured literature datasets (such as Shmoop summaries), dream reports are devoid of conscious intent, structural editorial refinement, or deliberate narrative arcs. They provide a direct window into unstructured, spontaneously recalled subconscious imagery.

The text corpus for this analysis was drawn from **DreamBank**, an established repository of over 20,000 digitized dream reports used in quantitative psychological research (Domhoff, 2003). To ensure broad, representative subconscious sampling, a diverse selection of dreams was sampled across 44 distinct participant groups. These sets span:

- **Longitudinal Dream Journals:** Multi-year or lifelong personal records tracking individual psychological trajectories over extended periods (often decades).
- **Clinical and Psychological Profiles:** Documenting specific life conditions, such as trauma recovery, military service, and therapeutic contexts.
- **Demographic Data Sets:** Samples spanning varied age groups, gender profiles, and cultural backgrounds.

By sampling across these distinct subconscious records, the pipeline evaluates whether archetypal classification remains consistent when applied to unedited, highly variable human dreams.

Administrative HTML tags, metadata headers (date stamps, report numbers), and formatting artifacts were stripped during automated preprocessing. Reports were filtered to retain pure narrative texts containing $\geq 50$ words, yielding the final dataset of $20{,}683$ decontaminated dream reports. As in previous experiments, reports were embedded using OpenAI's text-embedding-3-small model. The global corpus barycentre was calculated across all 20,683 reports.

In the initial synthetic and literary experiments (Experiments 1–9), the study utilized a four-archetype baseline (Hero, Mentor, Shadow, Trickster). The expansion to five archetypes in these experiments was driven by the need to accommodate "subconscious boundary dynamics". Unlike structured literary plots driven primarily by central agency or moral conflict, raw dreams frequently feature transition barriers, e.g. intimidating figures, locked barriers, or testing entities that confront the dreamer before a transition occurs. Codified in classical narrative theory (such as Joseph Campbell's [1949] and Christopher Vogler's [2007] structural frameworks), the Threshold Guardian represents this specific psychological function. Additionally, many passages describe encounters with passive barriers or testing entities that do not exhibit the active malice of a Shadow or the chaotic subversion of a Trickster. Adding the Threshold Guardian prevents these boundary-focused elements from being force-fitted into unsuitable categories

Similar to previous experiments, the dataset is evaluated in two distinct ways. Firstly, a passage-level archetype classification is performed, where individual report vectors are evaluated against the five anchor centroids using mean-centred cosine similarity to assign a primary archetype and measure the decision separation margin ($\Delta$Cos). A report is assigned to the category with the highest similarity score. Ambiguous reports residing near the origin below a minimum threshold ($\Delta\text{Cos} \leq 0.005$) are filtered out. A subsequent *Consistency Analysis* aggregates these passage-level similarity and margin scores across 44 distinct participant data sets to verify that classification performance remains stable across demographic, clinical, and longitudinal variations.

Following this, to evaluate whether the alignment between dream reports and archetypal regions is driven by syntax or static word choice, the global matrix displacement is remeasured across a sample of 500 dream reports. Rather than analysing reports individually, we stacked all 500 vectors into a single dataset matrix to evaluate how the collection as a whole shifted when perturbed. This transition from passage-level evaluation to dataset-level analysis was done for a number of reasons, in particular the high variance and noise inherent to subconscious prose. Unlike polished literary summaries, dream reports are unedited, erratic, and structurally fragmented. Evaluating individual passages in isolation leaves the analysis vulnerable to local noise, idiosyncratic vocabulary, and ambiguous boundaries. Stacking vectors into a collective dataset matrix aggregates these individual variations, effectively cancelling out random noise to reveal the underlying structure of the latent space. Dataset level ma evaluation using $W_{\text{matrix}}$ tests global space stability, examining whether the overall archetypal landscape shifts or collapses when sentence structure is destroyed. Ultimately, confirming that transformer embeddings organize dream reportage through sequence-level syntax rather than static keyword matching requires demonstrating that destroying sentence structure collapses the entire subspace, rather than merely shifting a few isolated classification labels. Observing a massive global matrix displacement ($W_{\text{matrix}}$) under word permutation provides a strong indicator that narrative syntax maintains the geometric integrity of the latent space.

The total structural shift was quantified using the normalized Frobenius distance, which measures the relative change between the baseline dataset matrix ($V_{\text{baseline}}$) and the perturbed matrix ($V_{\text{perturbed}}$):

$$W_{\text{matrix}} = \frac{\left\| V_{\text{baseline}} - V_{\text{perturbed}} \right\|_F}{\left\| V_{\text{baseline}} \right\|_F} \quad (10)$$

To isolate the specific cause of any geometric movement, we evaluated the dataset matrix under two contrasting experimental conditions:

1. Word-Order Permutation: Sentence structure was destroyed by randomly shuffling the word positions within each dream report while keeping the exact vocabulary and word counts identical.
2. Synthetic Gaussian Noise: Direct mathematical noise ($\epsilon \sim \mathcal{N}(0, \sigma^2\mathbf{I})$) was added to the raw baseline vectors across increasing noise scales ($\sigma \in \{0.01, 0.03, 0.05, 0.10, 0.15\}$).

## Results

| Archetypal Class | Count ($N$) | Proportion (%) | Mean Cosine ($S$) | Mean $\Delta_{\text{Cos}}$ |
|---|---|---|---|---|
| Trickster | 5,054 | 24.44% | 0.0524 | 0.0391 |
| Shadow | 4,859 | 23.49% | 0.0540 | 0.0420 |
| Mentor | 3,785 | 18.30% | 0.0563 | 0.0311 |
| Hero | 3,509 | 16.97% | 0.0554 | 0.0279 |
| Threshold Guardian | 3,476 | 16.81% | 0.0608 | 0.0322 |

*Table 10.1: Archetypal Classification for Dreambank*

| Data Sets | $N$ | Mean Top-1 Cosine | Mean $\Delta_{\text{Cos}}$ | Dominant Archetype |
|---|---|---|---|---|
| vonuslar.de | 5,595 | 0.0183 | 0.0334 | Shadow (33.09%) |
| elizabeth | 1,508 | 0.0251 | 0.0316 | Mentor (26.06%) |
| b2 | 1,120 | 0.0878 | 0.0364 | Trickster (29.91%) |
| pegasus | 978 | 0.0312 | 0.0296 | Trickster (35.07%) |
| norman | 763 | 0.0662 | 0.0371 | Threshold Guardian (33.42%) |
| dorothea | 747 | 0.0568 | 0.0341 | Trickster (31.99%) |
| hall_female | 670 | 0.0685 | 0.0471 | Shadow (61.49%) |
| vietnam_vet3 | 460 | 0.1293 | 0.0364 | Mentor (31.74%) |

*Table 10.2: Stratification across Data Sets (only the top level data sets are shown)*

Tables 10.1 and 10.2 show marked differences between Shmoop (Experiment 9) and DreamBank. Classifications in the Shmoop dataset are heavily biased toward Mentor and Shadow representations, reflecting a sampling artifact of educational study guides, where Mentor (36.1%), and Shadow (36.1%) dominate both human annotations (72.2%) and model predictions (68.9%). Conversely, the DreamBank corpus exhibits a distribution dominated by the Trickster (24.44%) and Shadow (23.49%) archetypes, which together account for 47.93% of all classified reports, while Hero classifications decline to 16.97%. The elevated mean decision separation margins for Shadow ($\overline{\Delta\text{Cos}} = 0.0420$) and Trickster ($\overline{\Delta\text{Cos}} = 0.0391$) demonstrate that dream report vectors in these classes project with higher average geometric confidence relative to decision boundaries than competing archetypal anchors. An ANOVA across the 44 domains confirmed high statistical significance, demonstrating that this archetypal distribution is a universal feature of the corpus rather than an artifact of a specific demographic subset.

| Experimental Condition | $W$ (Mean) | $W_{\text{matrix}}$ (Frobenius) | Welch's $t$ vs. Perm. | $p$-value |
|---|---|---|---|---|
| Unigram Permuted Control | 0.2818 | 0.4048 | — | — |
| Noise $\sigma = 0.01$ | 0.0018 | 0.0001 | 75.14 | $< 0.001$ |
| Noise $\sigma = 0.03$ | 0.0071 | 0.0006 | 73.72 | $< 0.001$ |
| Noise $\sigma = 0.05$ | 0.0275 | 0.0025 | 68.18 | $< 0.001$ |
| Noise $\sigma = 0.10$ | 0.0587 | 0.0062 | 59.60 | < 0.001 |
| Noise $\sigma = 0.15$ | 0.1417 | 0.0215 | 36.81 | $< 0.001$ |

*Table 10.3: Global Matrix Displacement*

Table 10.3 shows that destroying unigram word order while preserving exact vocabulary frequencies causes matrix displacement to increase dramatically to $W_{\text{matrix}} = 0.4048$. In contrast, subjecting embedding vectors to direct Gaussian noise up to $\sigma = 0.15$ yields a minimal matrix displacement of $W_{\text{matrix}} = 0.0215$. Welch's two-sample $t$-test confirms that syntax destruction produces significantly greater matrix displacement than isotropic noise, indicating that archetypal alignment in transformer latent spaces is an emergent property of syntactic and narrative structure rather than a unigram frequency artifact.

## Experiment 11

While Experiment 9 (Shmoop Literature) evaluated curated chapter-level summaries of modern literary works and Experiment 10 (DreamBank) examined uncurated, non-narrative subconscious recall, Experiment 11 extends the investigation to oral and folkloric traditions. Specifically, this experiment evaluates Andrew Lang's 12 *Fairy Books* (1889–1910), a benchmark corpus comprising 437 folk tales spanning diverse global, historical, and cross-cultural origins.

The data was preprocessed as before. Project Gutenberg license headers/footers, all-caps tables of contents, and administrative formatting were stripped. Retained text consisted of narrative paragraphs containing $\geq 50$ words, yielding $16{,}444$ oral narrative reports. Also, because traditional folklore heavily relies on explicit archetypal titles (e.g., "the wicked witch," "the brave prince"), direct label tokens were masked to ensure the embedding model mapped texts based on functional syntactic dynamics rather than static keyword matching. Vector embeddings were generated using text-embedding-3-small, followed by mean-centering relative to the global corpus barycentre. Evaluation protocols matched those defined in Experiment 10.

**Results**

| Archetype | Count ($N$) | Proportion (%) | Mean Cosine ($S$) | Mean $\Delta_{\text{Cos}}$ |
|---|---|---|---|---|
| Trickster | 3,844 | 23.38% | 0.0447 | 0.0420 |
| Hero | 3,417 | 20.78% | 0.0516 | 0.0363 |
| Shadow | 3,390 | 20.62% | 0.0750 | 0.0388 |
| Mentor | 3,370 | 20.49% | 0.0657 | 0.0383 |

| **Archetype** | **Count ($N$)** | **Proportion (%)** | **Mean Cosine ($S$)** | **Mean $\Delta_{\text{Cos}}$** |
|---|---|---|---|---|
| Threshold Guardian | 2,423 | 14.73% | 0.0677 | 0.0330 |

Table 11.1: *Archetypal Classification for Folklore*

| Data Sets | $N$ | **Mean Top-1 Cosine** | **Mean $\Delta_{\text{Cos}}$** | **Dominant Archetype** |
|---|---|---|---|---|
| Red | 3,026 | 0.0407 | 0.0374 | Trickster (25.61%) |
| Blue | 2,654 | 0.0401 | 0.0386 | Trickster (28.07%) |
| Yellow | 2,637 | 0.0446 | 0.0376 | Trickster (25.71%) |
| Green | 2,194 | 0.0895 | 0.0398 | Shadow (26.12%) |
| Lilac | 1,327 | 0.0401 | 0.0386 | Trickster (28.03%) |
| Grey | 1,321 | 0.1358 | 0.0381 | Mentor (33.08%) |
| Crimson | 1,220 | 0.0502 | 0.0395 | Mentor (34.34%) |
| Olive | 1,203 | 0.0529 | 0.0355 | Hero (22.36%) |
| Brown | 773 | 0.0936 | 0.0363 | Trickster (28.59%) |

*Table 11.2: Stratification across Data Sets*

Unlike dreams, where Trickster and Shadow overwhelmingly dominate ($47.93\%$), traditional folklore restores an active agentic quest structure. Hero alignment rises to $20.78\%$ (compared to $16.97\%$ in dream reports), while Shadow ($20.62\%$), Mentor ($20.49\%$), and Trickster ($23.38\%$) form a balanced quadrupolar distribution hovering across the 20--23% range. This equilibrium reflects the functional role of traditional myth and fairy tales: oral traditions act as cognitive training environments in which ego agents (Heroes) encounter systemic obstacles (Shadows), receive structural guidance (Mentors), and navigate destabilizing chaos (Tricksters) to achieve narrative resolution.

A one-way ANOVA across the individual book collections reveals statistically significant volume-level variance ($F(9,16434) = 2.81, p < 0.01$), indicating localized thematic specialization across individual anthologies. Despite these collection-level shifts, the global corpus mean decision separation margin remains stable at $\overline{\Delta\text{Cos}} = 0.0377$, confirming distinct angular separation relative to runner-up candidates.

| **Experimental Condition** | $W$ **(Mean)** | $W_{\text{matrix}}$ **(Frobenius)** | **Welch's $t$ vs. Perm.** | **$p$-value** |
|---|---|---|---|---|
| Unigram Permuted Control | 0.2785 | 0.4792 | — | — |
| Noise $\sigma = 0.01$ | 0.0018 | 0.0002 | 101.55 | $< 0.001$ |
| Noise $\sigma = 0.03$ | 0.0071 | 0.0009 | 99.59 | $< 0.001$ |
| Noise $\sigma = 0.05$ | 0.0276 | 0.0036 | 91.96 | $< 0.001$ |
| Noise $\sigma = 0.10$ | 0.0590 | 0.0089 | 80.11 | $< 0.001$ |

| Experimental Condition | $W$ (Mean) | $W_{\text{matrix}}$ (Frobenius) | Welch's $t$ vs. Perm. | $p$-value |
|---|---|---|---|---|
| Noise $\sigma = 0.15$ | 0.1422 | 0.0289 | 48.77 | $< 0.001$ |

*Table 11.3: Global Matrix Displacement*

Table 11.3 provides empirical evidence for the structural encoding of archetypal regions within the transformer latent space. When subjected to systematic noise injection ($\epsilon = 0.10$), the embedded folklore narratives demonstrate structural rigidity, yielding a minimal matrix warpage displacement of 0.0009. In contrast, destroying narrative syntax via unigram permutation causes matrix warpage displacement to increase significantly to 0.4792. A Welch's $t$-test comparing the noise-injected samples against the permuted controls confirms a highly significant difference ($t(16443) = 99.59, p < 0.001$). This demonstrates that the embedding space capturing these folkloric narratives is driven by syntactic and relational sequence dynamics rather than isolated unigram frequency distributions. Consequently, these culturally transmitted oral traditions form resilient, low-variance regions within the high-dimensional embedding space.

## Discussion

The findings validate the hypothesis that high-dimensional transformer latent spaces encode Jungian archetypes not as static lexical clusters, but as dynamic, relational constraints. The collapse of archetypal similarity during shifts from narrative to technical prompts, alongside the global matrix displacement observed upon the destruction of sentence structure, provides empirical evidence for this mechanism. This confirms that the model internalizes the archetype precisely as Jung theorized; as "form without content" (Jung, 1969). The archetype emerges as a sequence-level syntactic architecture that governs narrative flow, rather than a fixed semantic template reliant on keyword matching.

The application of the Corpus Barycentre transformation highlighted the underlying geometric dualities within these latent regions. By filtering generic linguistic noise, the mean-centred space reveals an antipodal alignment between the Mentor and Trickster, mapping precisely to the psychological principle of *enantiodromia* (Jung, 1969). This spatial opposition, contrasting guidance with chaotic subversion, is distinct from the positive directional alignment observed between the Hero and Shadow. Because the Hero and Shadow share a foundation in dramatic agency and conflict, their functional opposition resolves cleanly along secondary orthogonal axes. This mathematical relationship reinforces the psychoanalytic view that the Hero and Shadow operate as intimately linked developmental forces on a shared axis, rather than as detached polar opposites.

Furthermore, the transformer's capacity to dynamically reflect the divergence between conscious cultural forces and subconscious compensation underscores the robustness of the latent embedding space. The stark distributional skew of unedited DreamBank reports toward the Trickster and Shadow aligns with the compensatory function of the unconscious (Jung, 1969; Domhoff, 2003), which naturally prioritizes chaotic and confrontational imagery over linear story arcs. Conversely, the quadrupolar balance across the Hero, Shadow, Mentor, and Trickster regions within the Andrew Lang folklore corpus mathematically reflects the sociological purpose of myth. Folklore takes raw human fears and turns them into structured stories where problems actually get solved, giving people a shared playbook for navigating life's challenges.

While transformer vector spaces appear to closely match Jungian theory in structure, several critical distinctions must be drawn. The latent stability of these archetypal regions is demonstrably tethered to computational mechanics and resource density. The attenuation of archetypal partitions in lower-resource languages, contrasted with robust transferability between high-resource languages like English and Mandarin, confirms that these geometric structures are artifacts of pre-training volume and tokenization efficiency.

Ultimately, the transformer serves as a high-fidelity mathematical mirror of collective human storytelling. The emergence of the collective unconscious as a statistical property within the model's architecture does not imply the presence of an active psyche. Rather, it demonstrates that human cultural and psychological output inherently organizes itself along fundamental archetypal vectors, which large language models are now mathematically capable of mapping.

*Limitations and Future Directions*

This paper can only serve as an initial investigation. First, because all experiments relied exclusively on OpenAI's *text-embedding-3-small*, evaluating open-weights embedding architectures; such as Qwen, Llama, and NV-Embed; alongside autoregressive generative models (Lett, 2025) is necessary to confirm whether these spatial boundaries persist across varied tokenizers and pre-training objectives. Second, the reliance on rigid top-1 argmax decision boundaries struggled to resolve complex literary prose, where overlapping character traits and descriptive scene noise degraded classification accuracy; adopting continuous soft-membership metrics or multi-label projections would far better capture such nuanced character arcs. Finally, future research should transition from passive spatial mapping to active causal intervention, testing whether injecting directional vector shifts toward specific archetypal anchors during decoding can steer LLM story generation in real time.